\documentclass[nohyperref]{article}

\usepackage{microtype}
\usepackage{graphicx}

\usepackage{booktabs} 

\usepackage{hyperref}

 \usepackage[accepted]{icml2023}

\usepackage{amsmath}
\usepackage{amssymb}
\usepackage{mathtools}
\usepackage{amsthm}

\usepackage[capitalize,noabbrev]{cleveref}

\theoremstyle{plain}

\theoremstyle{definition}

\theoremstyle{remark}

\usepackage{graphicx}
\usepackage{float}
\usepackage{subfig}
\usepackage[textsize=tiny]{todonotes}

\icmltitlerunning{Subspace based Federated Unlearning}

\DeclareMathOperator*{\argmax}{arg\,max}
\DeclareMathOperator*{\argmin}{arg\,min}

\newcommand{\D}{\mathcal{D}}

\usepackage{bm}
\usepackage{multirow}

\begin{document}

\twocolumn[
\icmltitle{Subspace based Federated Unlearning 
}



\begin{icmlauthorlist}
\icmlauthor{Guanghao Li}{yyy}
\icmlauthor{Li Shen}{comp}
\icmlauthor{Yan Sun}{sch}
\icmlauthor{Yue Hu}{yyy}
\icmlauthor{Han Hu}{bit}
\icmlauthor{Dacheng Tao}{comp,sch}
\end{icmlauthorlist}

\icmlaffiliation{yyy}{National University of Defense Technology, China}
\icmlaffiliation{comp}{JD Explore Academy, China}
\icmlaffiliation{sch}{The University of Sydney, Australia}
\icmlaffiliation{bit}{Beijing Institute of Technology, China}

\icmlcorrespondingauthor{Li Shen}{mathshenli@gmail.com}

\icmlkeywords{Machine Learning, ICML}

\vskip 0.3in
]




\printAffiliationsAndNotice{}  
\begin{abstract}
Federated learning (FL) enables multiple clients to train a machine learning model collaboratively without exchanging their local data. 
Federated unlearning is an inverse FL process that aims to remove a specified target client’s contribution in FL to satisfy the user's right to be forgotten. Most existing federated unlearning algorithms require the server to store the history of the parameter updates, which is not applicable in scenarios where the server storage resource is constrained. In this paper, we propose a simple-yet-effective subspace based federated unlearning method, dubbed SFU, that lets the global model perform gradient ascent in the orthogonal space of input gradient spaces formed by other clients to eliminate the target client’s contribution without requiring additional storage. Specifically, the  server first collects the gradients generated from the target client after performing gradient ascent, and the input representation matrix  is computed locally by the remaining clients. We also design a differential privacy method to protect the privacy of  the representation matrix. Then the server merges those representation matrices to get the input gradient subspace and  updates the global model in the orthogonal subspace of the input gradient subspace to complete the forgetting task with minimal model performance degradation.
Experiments on MNIST, CIFAR10, and CIFAR100 show that SFU outperforms several state-of-the-art (SOTA)  federated unlearning algorithms by a large margin in various settings.

\end{abstract}


\section{Introduction}

The traditional training approach of deep learning usually brings together data from various participants. However, some data, e.g., medical records~\cite{liu2017detecting}, cannot be moved from  the hospital due to data privacy and personal-will concerns. Federated learning (FL)~\cite{mcmahan2017communication}  is one of the most popular decentralized machine learning methods to solve the above problems. It trains a global model by transmitting the model parameters between clients and the server, which obviates the need of transferring the underlying data~\cite{li2022fedhisyn,li2020review,li2020federated}.

Recent privacy legislations ~\cite{bussche2017eu, pardau2018california,shastri2019seven} provide data owners the right to be forgotten. In the machine learning context, the right to be forgotten requires that (i) the user data is deleted from the entity storing it and (ii) the influence of the data on the model is removed~\cite{halimi2022federated}. Federated unlearning is the embodiment of the user's right to be forgotten in the FL scenario, where the goal  is to remove the contribution of specific clients' data from the global model while maintaining the model's accuracy. Three challenges in FL make the traditional machine unlearning approach unsuitable for federated unlearning: (1) Limited Access to the dataset:  The server in FL cannot directly access all data and perform related operations, which makes forgetting techniques that rely on dataset segmentation cannot be applied to FL scenarios. (2) Model aggregation: The initial model of each client in each training round depends on the model aggregation of the clients that participated in the training in the previous round. Removing the contribution of a client will affect the model aggregation and affect the subsequent training of other clients~\cite{nasr2018comprehensive,melis2019exploiting,song2020analyzing}. (3) Client selection: Since not all clients can participate in the training process in each round, the contribution of each client to the global model is not continuous, and the contribution can only be made when the client is selected.

There has been some current work on federated unlearning.  To remove the contribution of a specific target client in the final global model, a naive idea is to retrain the model on the remaining clients after removing the target one~\cite{bourtoule2021machine}. However, some clients  may only have limited storage, and they may delete data at any time after the training process. It will make even the most naive way of retraining the model from scratch impractical since the client may not have the same data as that used in the training time. Another idea in federated unlearning is to store the client's historical updated gradient data in the server and use it to roll back the trained global model~\cite{wu2022federated}. {The above unlearning methods all require the client or server to retain some additional data or gradient information, which is not possible in FL scenarios with limited storage resources.}

In each FL training round, local training of each client is a process that reduces the empirical loss~\cite{rumelhart1986learning}. We argue that unlearning can be formulated as the inverse process of learning, in the sense that the gradient ascent on the target client can realize the forgetting of the client data. However, the loss is unbounded and we need to limit the gradient of training to ensure the quality of the model after unlearning~\cite{chen2015fast}. The whole process can be regarded as a constraint-solving problem to maximize the empirical loss of the target client within the constraints of the model performance. In this paper, we propose a Subspace-based Federated Unlearning method, dubbed SFU.  SFU restricts the gradient of the target client's gradient ascent to the orthogonal space of the input space of the remaining clients to remove  the target client's contribution from the final trained global model. In SFU, the server only needs the gradient information provided by the target client and the representation matrix information provided by other clients, without directly accessing the original data of each client. On the other hand, SFU can be used for models in any training phase without considering the specific details of model training and model aggregation. At the same time, SFU does not require the client or server to store additional historical gradient information or data.

Specifically, SFU participants can be divided into three kinds of roles: the target client to be forgotten, the remaining clients, and the server. In SFU, the target client performs gradient ascent locally based on the global model and sends the gradient to the server; each remaining client selects a certain amount of local data to build a representation matrix and send it to the server; the server receives the representation matrix from each client and merges it to obtain the input subspace with  Singular Value Decomposition (SVD)~\cite{hoecker1996svd}; the server finally projects the gradient of the target client into the orthogonal subspace of the input space and updates the global model with it. In addition, we design a differential privacy method to protect the privacy of clients in the process of sending the representation matrices~\cite{li2021survey,truex2019hybrid}.  It needs each client to add random perturbation factors to each vector of the representation matrix  to prevent possible privacy leaks and those perturbation factors have no effect on the  input space search and the final model. Empirical results show that SFU beats other SOTA baselines with 1\%-10\% improvement in test sets.

In the end, we summarize the main contributions as follows:
\begin{itemize}
\item  We first introduce subspace learning to federated unlearning and propose a new federated unlearning algorithm called SFU, which trains the global model by gradient ascent in an orthogonal subspace perpendicular to the input space of the remaining clients to achieve the forgetting goal  without large performance loss and additional storage costs.

\item We design a differential privacy method to prevent the possible privacy leakage caused by the transmission of the client representation matrix during the process of unlearning. This method adds random perturbation factors to each vector of the representation matrix but does not affect the unlearning process.

\item  We conduct extensive experiments to evaluate the effectiveness of SFU, which significantly outperforms several SOTA baselines on various datasets including MNIST, CIFAR10, and CIFAR100.

\end{itemize}


\section{Related Work}
\textbf{Machine unlearning.}
The term ``machine unlearning"  is a process to forget a piece of training data completely which needs to revert the effects of the data on the extracted features and models. Machine unlearning is first proposed by ~\citet{cao2015towards}, and they transform the statistical query learning into a summation form and achieve unlearning by updating a small part of the summation. However, this algorithm only works for those transformable traditional machine learning methods, machine unlearning for different ML models has been explored. ~\citet{ginart2019making} formulate the problem and the concept of effective data deletion in machine learning. They also propose two efficient deletion solutions for the K-means clustering algorithm. ~\citet{izzo2021approximate} focus on supervised linear regression and propose an approximate data deletion method called projective residual update (PRU) for linear and logistic models. The computational cost of PRU  is linear in the feature dimension, but PRU is not applicable for more complex models such as neural networks. ~\citet{baumhauer2022machine} introduce the more general framework SISA to reduce the computational overhead associated with forgetting. The main idea of SISA is to split the training data into several disjoint fragments, each of which trains a sub-model. To remove a specific sample, the algorithm simply retrains the sub-model that is learned from this sample. However, existing machine learning works focus on ML models in traditional centralized settings, where training data is assumed to be globally accessible, which is not suitable for learning in FL settings.

\textbf{Federated unlearning.}
Compared with centralized learning, there are few works on unlearning in FL. ~\citet{liu2021federaser} first introduce unlearning into the field of FL and propose FedEraser. The main idea is to adjust the historical parameter updates of federated clients through the retraining process in FL and reconstruct the unlearning model. However, this process requires additional communication between the client and the server. Recently, ~\citet{wu2022federated} develop a forgetting approach that removes historical parameter updates from the target client and recovers model performance through a knowledge distillation process. Both of these steps require the server to keep a history of all client updates. In addition, the knowledge distillation approach requires the server to have some additional unlabeled data. In some application scenarios, meeting these requirements may not be a matter of course.  In contrast, our method does not need the server to store historical updates or additional unlabeled data  and  has better privacy. Our method mainly solves the case where a client (referred to as the target client) wants to  remove its contribution from the global model. We let the global model perform gradient ascent in the orthogonal subspace of the input space of remaining  clients to achieve federated unlearning.  Different from these works (and the one we propose), \citet{wang2022federated} propose a forgetting framework for forgetting a specific class or category in FL.

Our approach is closely related to the federated unlearning approach recently proposed by ~\citet{halimi2022federated}. They formulate the unlearning problem as a constrained maximization problem by restricting to an $\ell_2$-norm ball around a suitably chosen reference model and allowed the target client to perform the unlearning by using the Projected Gradient Descent (PGD) algorithm~\cite{thudi2022unrolling}. However, $\ell_2$-norm ball can not provide an effective guarantee for the performance of the model after unlearning. We add constraints on the performance of the unlearning model. Our method restricts the gradient ascent to a subspace orthogonal to the input space of other clients, and this constrained gradient update has minimal impact on the performance of the model on other clients. We will quantitatively show that our method has a performance guarantee.
\section{Methodology}\label{methodology}

We propose a novel federated unlearning method, as shown in Algorithm~\ref{alg:SFU}, that can eliminate the client’s contribution and vastly reduce the unlearning cost in the FL system. This method does not require the server to keep the history of parameter updates from each client or additional training. The key idea is to use the restricted gradient information of the target client to modify the final trained model.

\begin{algorithm}[tb]
   \caption{Subspace-based Federated Unlearning (SFU)}\label{SFU}
   \label{alg:SFU}
\begin{algorithmic}[1]
   \STATE {\bfseries Input:} The number of samples in each client $p$, the global model  $w$, local dataset $\D^i$ of  client $i$,  random factors $\lambda_i^l$ for layer $l$ in client $i$.
   
   \STATE \textbf{Target client $C_I$:}
   \STATE  $g_i \gets -\eta\ell_i(w; (x, y))$;
  
   \STATE  Send $g_i$ to the server;
   \STATE \textbf{Other clients:}
   
\STATE Representation  matrix for layer $l$ in client $i$: $\bm{R}_i^l =  [\lambda_i^l x_{1}^l, \lambda_i^l x_{2}^l, ..., \lambda_i^l x_{p}^l ]$;

 \STATE Send $\bm{R}_i^l $ to the server;

\STATE \textbf{server:} 

\STATE$\bm{R}^l = [\bm{R}_1^l, ..., \bm{R}_{C_I+1}^l, \bm{R}_{C_I+1}^l ..., \bm{R}_N^l] $;

\STATE Performing SVD on $\bm{R}^l =\bm{U}^l\bm{\Sigma}^l(\bm{V}^l)^T$; 

 \STATE $S^l=span\{\bm{u}_{1,1}^l,\bm{u}_{2,1}^l, ...,\bm{u}_{k,1}^l\}$;
 
\STATE  $\tilde{g_{C_I}}$ = $proj(g_{C_I} , S)$;
 
\STATE $w = w - (g_{C_I}-\tilde{g_{C_I}})$
   
\end{algorithmic}
\end{algorithm}

\subsection{Problem Setup}

Suppose that there are $N$ clients, denoted as $C_1, ..., C_N$,  respectively. Client $C_i$ has a local dataset $\D^i$. The goal of traditional FL is to collaboratively learn a machine learning model $w$ over the dataset $\D\triangleq \bigcup_{i\in[N]}\D^i$ :
\begin{equation}\label{FL-formulation}
    \argmin_{w} \mathcal{L}(w) = \sum_{i=1}^N \frac{|\D^i|}{|\D|}L_i(w),
\end{equation}
\begin{equation}\label{FL-formulation}
    w^* = \argmin_{w} \mathcal{L}(w),
\end{equation}
where $L_i(w) = \mathbb{E}_{(x,y)\sim \D^i} [\ell_i(w; (x, y))]$ is the empirical loss of $C_i$ and during federated training, each client  minimizes  their empirical risk $L_i(w)$, $w^*$  is the final model trained by the FL process.

Now we consider how to forget the contribution of the target client $C_I$. A natural idea  is  to increase  the empirical risk $L_{C_I}(w)$ of the target client $C_I$, which is equivalent to reversing the learning process. However, simply maximizing the loss  can influence the effect of the model on other clients. Federated unlearning needs to forget the contribution of the target client $C_I$ while ensuring the overall model performance. Thus, the objective of federated unlearning is defined below:

\begin{equation}\label{FLUL-formulation}
\begin{aligned}
   & \argmax_{w} L_i(w) = \mathbb{E}_{(x,y)\sim \D^i} [\ell_i(w; (x, y))]\\
   &s.t. \qquad \mathcal{L}^{ul}(w) - \mathcal{L}^{ul}(w^*) \leq \delta \\
\end{aligned}
\end{equation}

where $\delta$  is a small change in the empirical loss, $\mathcal{L}^{ul}()$ is the empirical loss of the FL system after removing the  target client. 
\begin{equation}
    \mathcal{L}^{ul}(w) = \sum_{i\in[N\backslash C_I]} \frac{|\D^i|}{|\D^{un}|}L_i(w),
\end{equation}
where $\D^{un}\triangleq \bigcup_{i\in[N\backslash C_I]}\D^i$  is the remaining data set after removing the target client.

\subsection{Unlearning Metrics}
{Comparing the difference between the unlearned model and the retrained model is one of the criteria used to measure the effect of unlearning. Common dissimilarity metrics include model test accuracy difference~\cite{bourtoule2021machine}, $\ell_2$-distance~\cite{wu2020deltagrad} or, and Kullback-Leibler (KL) divergence~\cite{sekhari2021remember}. However, in the FL scenario, it is difficult to intuitively indicate whether the contribution of a given client can be removed from the evaluation method based on model differences. Other metrics include privacy leakage in the differential privacy framework ~\cite{sekhari2021remember} and membership inference attacks \cite{graves2021amnesiac,baumhauer2022machine}. In this paper, we use the backdoor triggers~\cite{gu2017badnets} as an effective way to evaluate the performance of unlearning methods, similar to ~\citet{wu2022federated}. In particular, the target client uses a dataset with a certain fraction of images that have a backdoor trigger inserted in them. Thus, the global FL model becomes susceptible to the backdoor trigger. Then, a successful unlearning process should produce a model that reduces the accuracy of the images with the backdoor trigger, while maintaining good performance on regular (clean) images. Note that we use the backdoor triggers as a way to evaluate the performance of unlearning methods; we do not consider any malicious client~\cite{xie2020dba,bagdasaryan2020backdoor,fung2020limitations}

\subsection{Subspace-based Federated Unlearning (SFU)}


{We introduce a novel Subspace-based federated unlearning framework, named SFU. The main insight of the SFU is that we constrain the gradients generated by the target client's gradient ascent to the input subspace of the other clients to remove the contribution of the target client from the global model. As shown in Fig.~\ref{fig:pipeline}, SFU participants can be divided into three kinds of roles: the target client to be forgotten, the remaining clients, and the server. The target client performs gradient ascent to upload the gradient to the server. Other clients compute the local representation matrix and upload it to the server. The server is responsible for the computation of other client input Spaces and the unlearning update of the global model. Next, we will introduce the specific tasks of the three participants respectively.}

\subsubsection{ Gradient ascent on the target client}
To find a model with a large empirical loss in target client $C_I$, we can simply make several local passes of (mini-batch stochastic) gradient ascent in client $C_I$ and add these gradient updates to the global model. In order to satisfy the constraints of Eq.~(\ref{FLUL-formulation}), we need to consider a more reasonable way to add the updated gradient of client $C_I$ to the global model~\cite{zhou2020bypassing,qian2015efficient}.

Given a neural network $W$ and an input $\textbf{x}$ we can obtain an output $\textbf{y}$:
\begin{equation}
    W\textbf{x} =  \textbf{y}_1.
\end{equation}
When this model accepts a gradient update $\Delta w$, the output becomes:
\begin{equation}
    (W + \Delta w) \textbf{x} =  \textbf{y}_2.
\end{equation}
The difference between the two outputs is:
\begin{equation}
\label{or}
    \Delta \textbf{y} =  \textbf{y}_2 - \textbf{y}_1 = (W + \Delta w) \textbf{x} - W\textbf{x} = \Delta w \textbf{x}.
\end{equation}
When $ \Delta \textbf{y} $  is 0, the difference between the two outputs is minimized, which requires the updated gradient $\Delta w$ to be perpendicular to the original input gradient subspace  $\textbf{x}$. Therefore, we can project the updated gradient of the target client $C_I$ into the orthogonal space of the gradient subspace of $\D^{un}$ to minimize the degradation of the glob model performance~\cite{farajtabar2020orthogonal}.

\subsubsection{Computation of representation matrix}

We first need to consider how to represent the input space in $\D^{un}$, the data  of other clients. For an individual network, we construct the gradient subspace by the following two steps:

\begin{itemize}
\item For each layer $l$ of the network, We  first construct a representation matrix, $\bm{R}^l =[\textbf{x}_{1}^l, \textbf{x}_{2}^l, ..., \textbf{x}_{n_s}^l ]$ concatenating $n_s$ representations along the column obtained from forward pass  of $n_s$ random samples from the current training dataset through the network.
\item Next, we perform SVD on $\bm{R}^l =\bm{U}^l\bm{\Sigma}^l(\bm{V}^l)^T$ followed by its $k$-rank approximation $(\bm{R}_1^l)_k$ according to the following criteria for the given coefficient, $\epsilon^l$ :
\begin{equation}
    ||(\bm{R}^l))_k||_F^2 \geq \epsilon^l||\bm{R}^l||_F^2.
\end{equation}
$S^l=span\{\bm{u}_{1,1}^l,\bm{u}_{2,1}^l, ...,\bm{u}_{k,1}^l\}$, spanned by the first $k$ vectors in $\bm{U}_1^l$ as the space of significant representation at layer $l$ since it contains all the directions with highest singular values in the representation~\cite{saha2021gradient}.
\end{itemize}
For FL scenarios, we need the data on each client to seek the gradient subspace of the $\D^{un}$. First,  all clients  excluding the target client $C_I$ select the same number of  representations matrix of local samples for each layer $\bm{R}_i^l$ and send them to the central server to construct the representation matrix. 

To protect the privacy of the representation matrix, we design a differential privacy algorithm. We add random factors $\lambda_i^l$ to the representation of layer $l$ from  client $i$ to avoid leaking data information about the client data and it does not affect the search process of the subspace because of the nature of the orthogonal matrix.In Eq.~(\ref{or}), if 
\begin{equation}
    \Delta w \textbf{x} = 0,
\end{equation}
we have
\begin{equation}
    \Delta w (\lambda \textbf{x}) = 0.
\end{equation}

 The final set of representation matrix in the server is $\bm{R} = \left\{\bm{R}^1, \bm{R}^2, ..., \bm{R}^L\right\} $, and $\bm{R}^l =[\lambda_{1}^l \textbf{x}_{1,1}^l, \lambda_{2}^l \textbf{x}_{2}^l, ..., \lambda_{N}^l \textbf{x}_{n_s}^l ]$.

\subsubsection{Update of the global model on the server}
After several local passes of (mini-batch stochastic) gradient ascent, client $C_I$ sends the updated gradient $g_{C_I}$ to the server. The server performs the update of the global model $w$ after collecting the set of representation matrix $\bm{R}$ and the  gradient $g_{C_I}$. The server first perform SVD~\cite{rumelhart1986learning} on $\bm{R}$ to get the set of input gradient subspace $S = \left\{S^1, S^2, ..., S^L\right\} $. To achieve the goal of Equation~\ref{FLUL-formulation}, we need to project $g_{C_I}$ onto $S$  and get projection $\tilde{g_{C_I}}$. $g_{C_I}-\tilde{g_{C_I}}$ orthogonal to $\bm{R}$  and the server update the global model $w$ with $g_{C_I}-\tilde{g_{C_I}}$:
\begin{equation}\label{ct1}
    w = w - (g_{C_I}-\tilde{g_{C_I}}).
\end{equation}

The updated model $w$ removes the contribution of the target client ${C_I}$ and maintains a similar performance to the original global model.

\begin{figure}
    \centering
    \setlength{\abovecaptionskip}{0.cm}
    \includegraphics[width=0.5\textwidth]{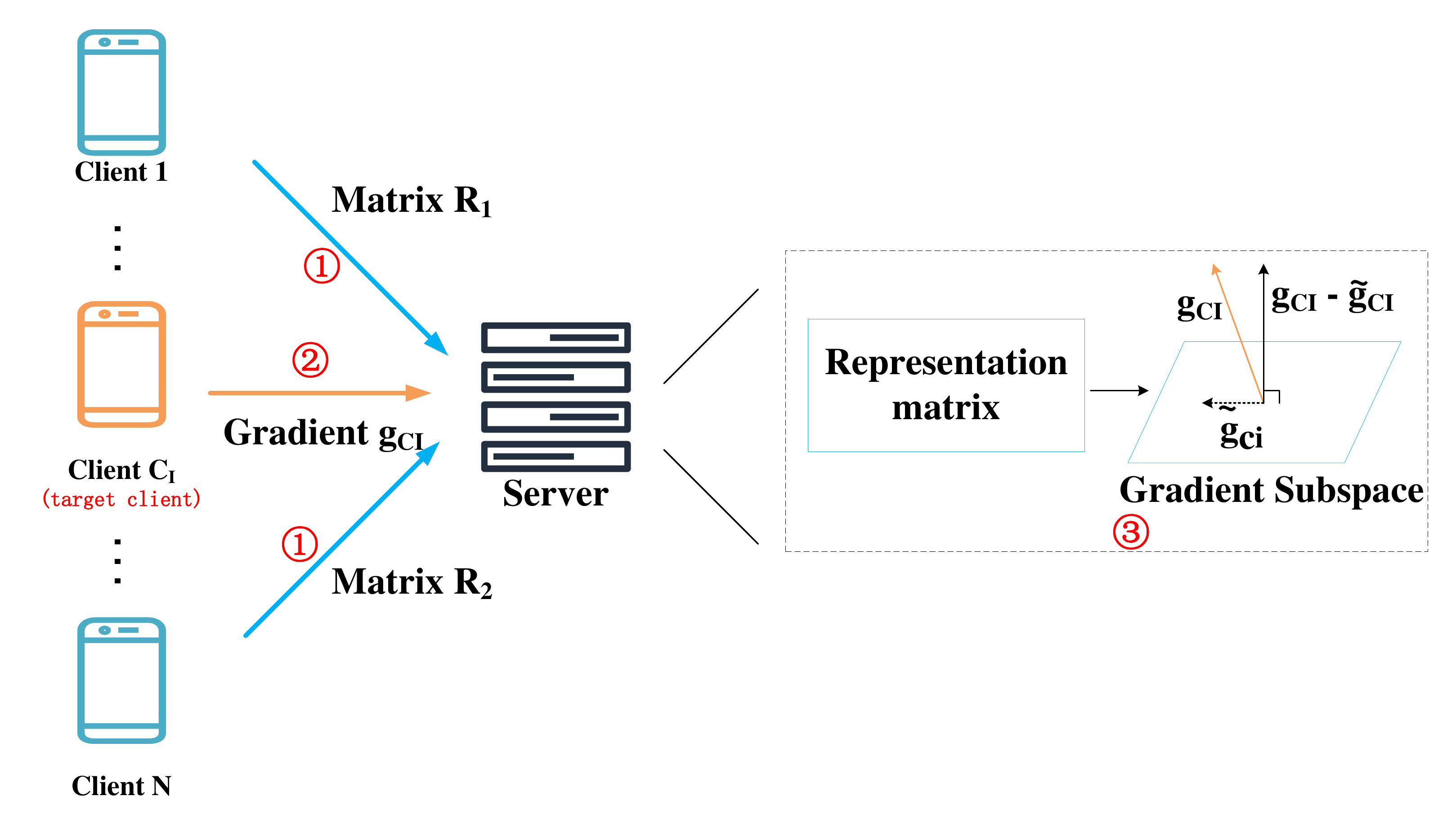}
    \caption{The pipeline of the SFU.The whole process takes place after the FL model has been trained. The orange client represents the target client whose contribution is to be removed; The blue ones represent other clients. The boxes on the right of the image represent global model updates that happen on the server. }
    \label{fig:pipeline}
\vspace{-0.5cm}
\end{figure}

After a global model is trained, a complete SFU training process mainly includes three steps as shown in Fig.1:
\begin{itemize}

\item \textbf{Step 1.}\ Besides target client ${C_I}$, each client selects the same number of samples to calculate the representation matrix $\bm{R}^1$ for each  layer$l$ of the network  and sends it to the server after adding random factors $\lambda_i^l$.

\item  \textbf{Step 2.}\ The target client ${C_I}$ performs  several local passes of  gradient  ascent locally and sends the updated gradient to the server.

\item  \textbf{Step 3.}\ The server perform SVD on the set of representation $\bm{R}$ to get the set of input gradient subspace $S$,  project $g_{C_I}$ onto $S$ and removes the contribution of the target client ${C_I}$ by updating the global model $w$ as Eq.~\ref{ct1}.
\end{itemize}

In the end, we give several comments on the proposed SFU algorithm. Note that subspace learning has been commonly used to solve continual learning \cite{saha2021gradient}, meta learning \cite{jiang2022subspace}, adversarial training \cite{li2022subspace}, and fast training of deep learning models \cite{li2022low}. However, SFU is the first work to use the orthogonal space of input gradient space for federated unlearning. In addition, SFU needs to seek subspace from dispersed stored data and should consider the privacy leakage risk.

\section{Experiments}

\begin{table*}
\centering
\footnotesize
\caption{Accuracy results after unlearning (IID).}
\label{tabatk}
\vspace{-0.2cm}
\begin{tabular}{|c|c|c|c|c|c|c|c|c|c|c|c|c|c|}
\hline

&&\multicolumn{2}{c|}{{FedAvg}}&\multicolumn{2}{c|}{{UL}}&\multicolumn{2}{c|}{{GA}}&\multicolumn{2}{c|}{{SFU}}\\
\hline

{Dataset}&{network}&{ test acc}&{atk acc}&{ test acc}&{atk acc}&{ test acc}&{atk acc}&{ test acc}&{atk acc}\\
\hline
\multirow{2}{*}{MNIST}  & {MLP}& 97.73 & 93.26 & 77.19 & 0.0 & 96.80 & 0.0 & \textbf{97.66} & 0.15 
  \\  \cline{2-2}

&{CNN} &  99.31 & 91.29 & 42.36  & 0.0 & 92.33 & 19.28 & \textbf{99.29} & 0.21 \\  \cline{2-2}

\hline
\multirow{2}{*}{CIFAR10}  & {MLP} &  49.15 & 89.66 & 26.17 & 0.0 & 48.61 & 0.01 & \textbf{49.08} & 0.0   \\  \cline{2-2}

&{CNN} &  72.83 & 99.36  & 18.75 & 0.0 & 57.48 & 0.37 & \textbf{57.75}  & 0.0\\  \cline{2-2}
&
{ResNet} &  76.12 & 98.27 & {43.75} & 0.0 & \textbf{73.58} & 65.98 & {44.60} & 0.0   \\  \cline{2-2}
\hline
\multirow{2}{*}{CIFAR100}  & {MLP} & 18.86 & 9.93 & 2.24 & 0.0 & 13.42 & 0.00 & \textbf{18.70 }& 0.0  \\  \cline{2-2}

&{CNN} &  37.47 & 97.77 & 2.51 & 0.0 & 20.12 & 1.41 & \textbf{36.31} & 0.0   \\  \cline{2-2}
&
{ResNet} &  39.68 & 90.38 & 7.38 & 0.0 & \textbf{37.63} & 6.73 & {26.84} & 0.0   \\  \cline{2-2}
\hline
\end{tabular}
\end{table*}

\begin{table*}
\centering
\footnotesize
\caption{Accuracy results after retraining (IID).}
\vspace{-0.2cm}
\label{tabacc}
\begin{tabular}{|c|c|c|c|c|c|c|c|c|c|c|c|c|c|}
\hline

&&\multicolumn{2}{c|}{{Retraining}}&\multicolumn{2}{c|}{{UL-Distillation}}&\multicolumn{2}{c|}{{GA-retraining}}&\multicolumn{2}{c|}{{SFU-retraining}}\\
\hline

{Dataset}&{network}&{ test acc}&{atk acc}&{ test acc}&{atk acc}&{ test acc}&{atk acc}&{ test acc}&{atk acc}\\
\hline
\multirow{2}{*}{MNIST}  & {MLP}& 97.88 & 0.0 & 97.88 & 0.0 & 97.05 & 9.30 & \textbf{97.91} & 0.28 
  \\  \cline{2-2}

&{CNN} &  99.48 & 11.55 & 99.20  & 0.31 & 99.34 & 37.13 & \textbf{99.37} & 12.63  \\  \cline{2-2}

\hline
\multirow{2}{*}{CIFAR10}  & {MLP} &  48.77 & 0.00 & 47.83 & 0.0 & {48.09} & 20.36 & \textbf{48.81} & 0.10   \\  \cline{2-2}

&{CNN} &  74.86 & 0.00 & 72.33 & 0.51 & 72.43 & 26.47 & \textbf{72.87} & 2.18  \\  \cline{2-2}
&
{ResNet} &  76.95 & 7.23 & \textbf{77.47} & 9.68 & {77.19} & 81.06 & 76.84 & 3.12   \\  \cline{2-2}
\hline
\multirow{2}{*}{CIFAR100}  & {MLP} & 18.47 & 0.0 & 9.43 & 0.0 & {18.81} & 2.46 & \textbf{18.46 }& 0.0  \\  \cline{2-2}

&{CNN} &  38.25 & 0.0 & 27.05 & 0.0 & 37.08 & 51.36 & \textbf{37.81} & 14.39   \\  \cline{2-2}
&
{ResNet} &  40.15 & 0.81 & 38.43 & 0.79 & 38.11 & 61.91 & \textbf{40.93} & 0.92   \\  \cline{2-2}
\hline
\end{tabular}
\vspace{-0.4cm}
\end{table*}

In this section, we empirically evaluate the  effectiveness of SFU using different model architectures on three datasets. We divide the unlearning process into two parts: the removal of specific client contributions and the recovery of model performance. The experimental results show that our unlearning strategies can effectively remove the contribution of the target client from the global model with low-performance loss and can quickly recover the accuracy of the model in a few rounds of training. We first introduce the experimental setup and then present the  evaluation results.

\subsection{Experimental Setup}

\textbf{Datasets:} 
We evaluate SFU using three popular datasets: MNIST\cite{xiao2017fashion}, CIFAR10, and CIFAR100~\cite{krizhevsky2009learning} as described below: 
		\begin{itemize}
			\item {MNIST:} It is a dataset that contains a training set of 60,000 examples and a test set of 10,000 examples. Each example is a 28×28 grayscale image associated with a label	from 10 classes.

			\item {CIFAR10:} It  consists of 60,000 32 × 32 color images in 10 classes, with 6000 images per class. There are 50,000 training images and 10,000 test images.
			\item {CIFAR100:} It has 100 classes. Each class has 600 color images of size 32 × 32, of which 5000 are used as a training set, and 100 are used as a test set.

		\end{itemize}
The training difficulty of the dataset is increasing from MNIST, CIFAR10 to CIFAR100. We adopt two data distribution Settings, including IID ({I}ndependent-{I}dentically-{D}istributed) as well as Non-IID ({N}on {I}ndependent-{I}dentically-{D}istributed). The Non-IID setting that we adopt is Dirichlet($\mathcal{\beta}$): Label distribution on each device follows the Dirichlet distribution, where $\mathcal{\beta}$ is a concentration parameter ($\mathcal{\beta} > 0$). 

\textbf{Baselines.}  We chose three typical federated unlearning algorithms as our baseline:

(1) Retraining: This method retrains the entire FL system without the target client being forgotten, which is an effective but computationally and communicationally expensive algorithm;

(2) ``UL"~\citet{wu2022federated}: This method  forgets the target client by  subtracting  historical parameter updates  of the target client from the global model.  Then, ``UL"  uses the knowledge distillation method to remedy the skew of the unlearning model caused by the subtraction without using any data from the target clients. 

(3)  ``GA": GA uses gradient ascent information on the target client to update the global model. Gradient ascent makes the loss of the model increase, which is the inverse process of learning and can achieve unlearning. However, the loss of gradient ascent is unbounded so we set the gradient clip norm when the global model is updated to reduce the probability of producing a random model.

\begin{figure*}[ht]
 \footnotesize
\centering \vskip -2pt
\centering
\setlength{\abovecaptionskip}{0.cm}
\subfloat[MNIST]{\includegraphics[width=0.3\linewidth]{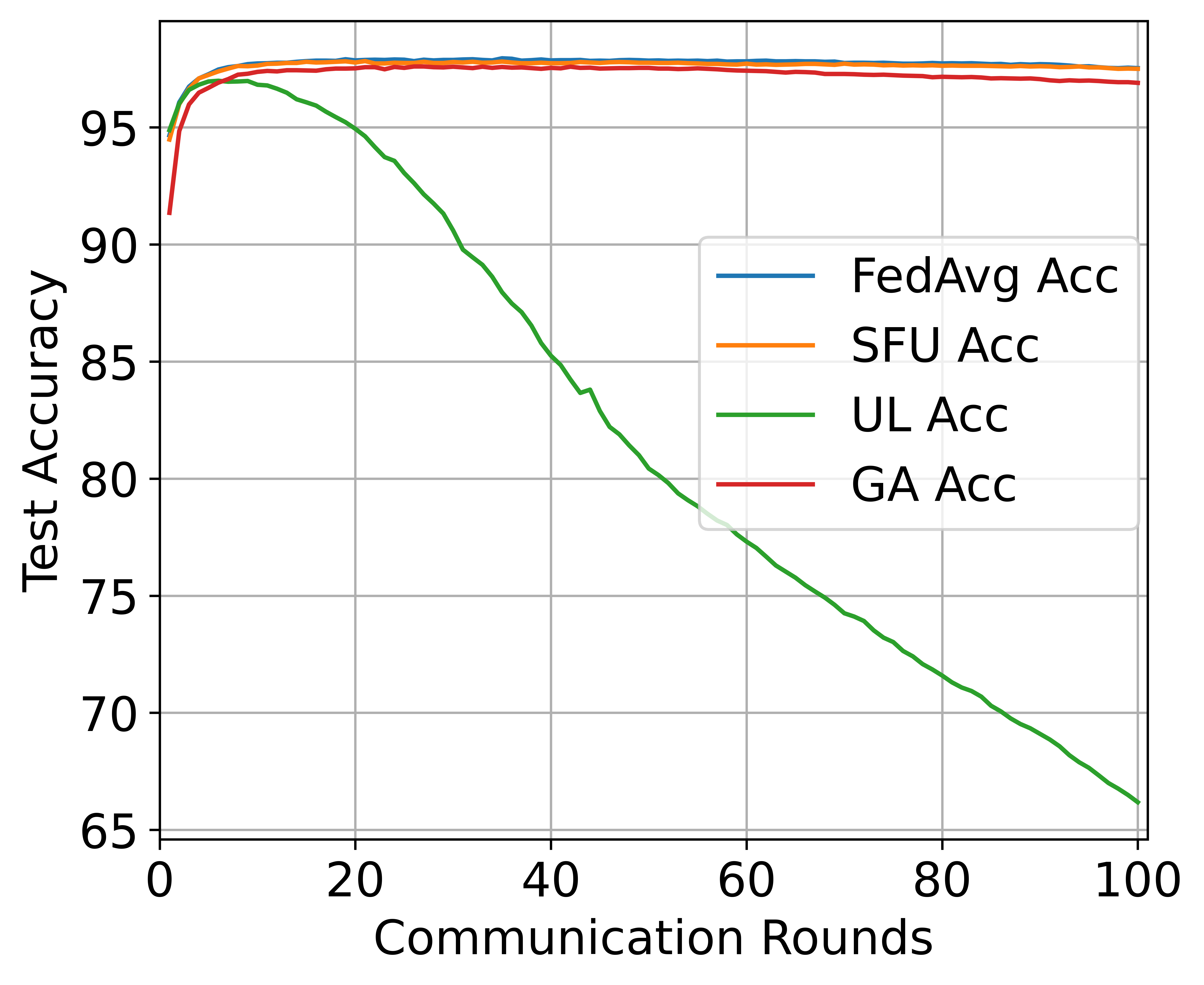}%
}
\hfill
\subfloat[CIFAR10]{\includegraphics[width=0.3\linewidth]{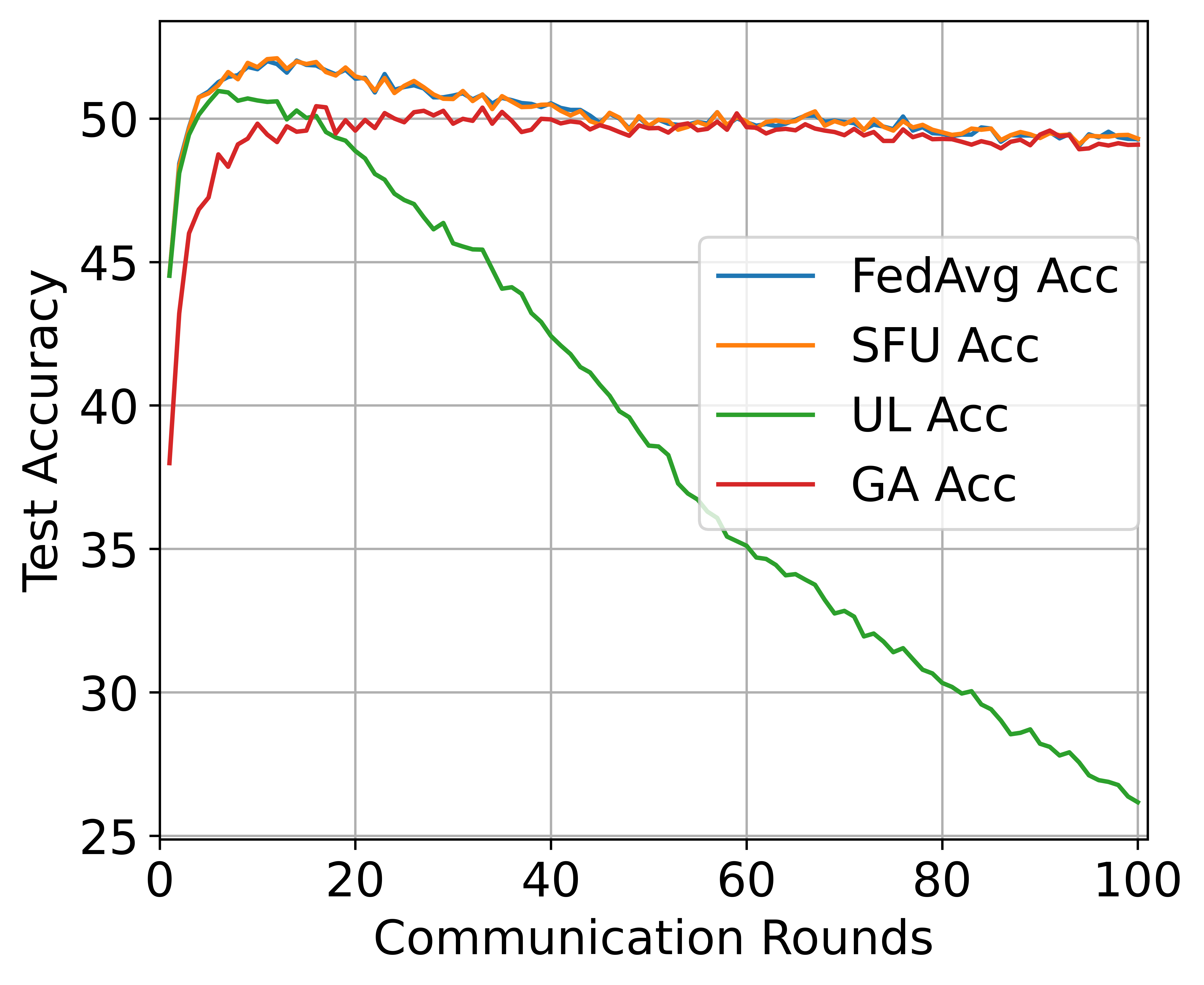}%
}
\hfill
\subfloat[CIFAR100]{\includegraphics[width=0.3\linewidth]{figure/cifar10_mlp.png}}
\caption{Model accuracy after execution of SFU and other baselines at various stages of FL model training. }
\label{fig:datadis}
\vspace{-0.4cm}
%
\centering
\setlength{\abovecaptionskip}{0.cm}
\subfloat[MNIST]{\includegraphics[width=0.3\linewidth]{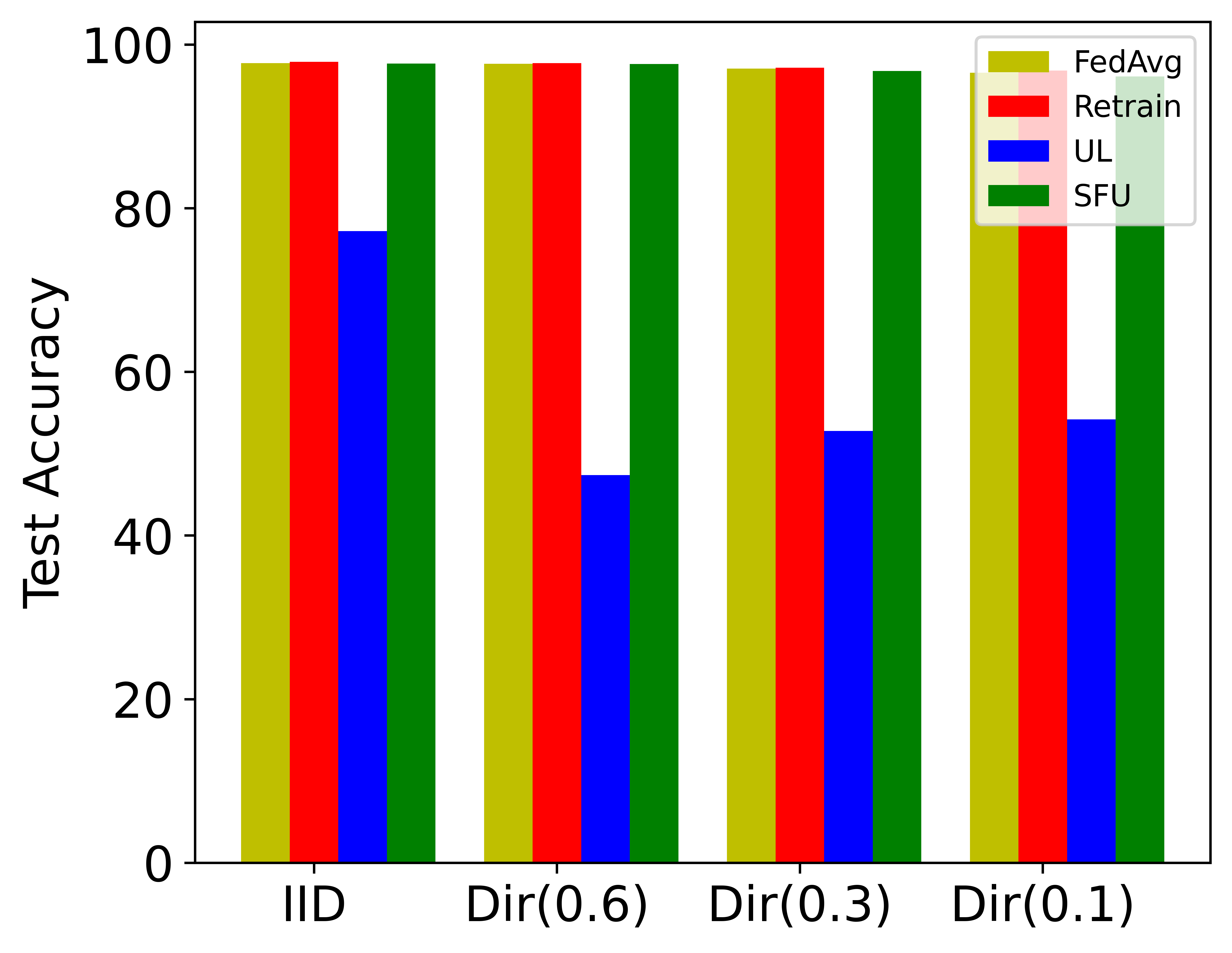}%
}
\hfill
\subfloat[CIFAR10]{\includegraphics[width=0.3\linewidth]{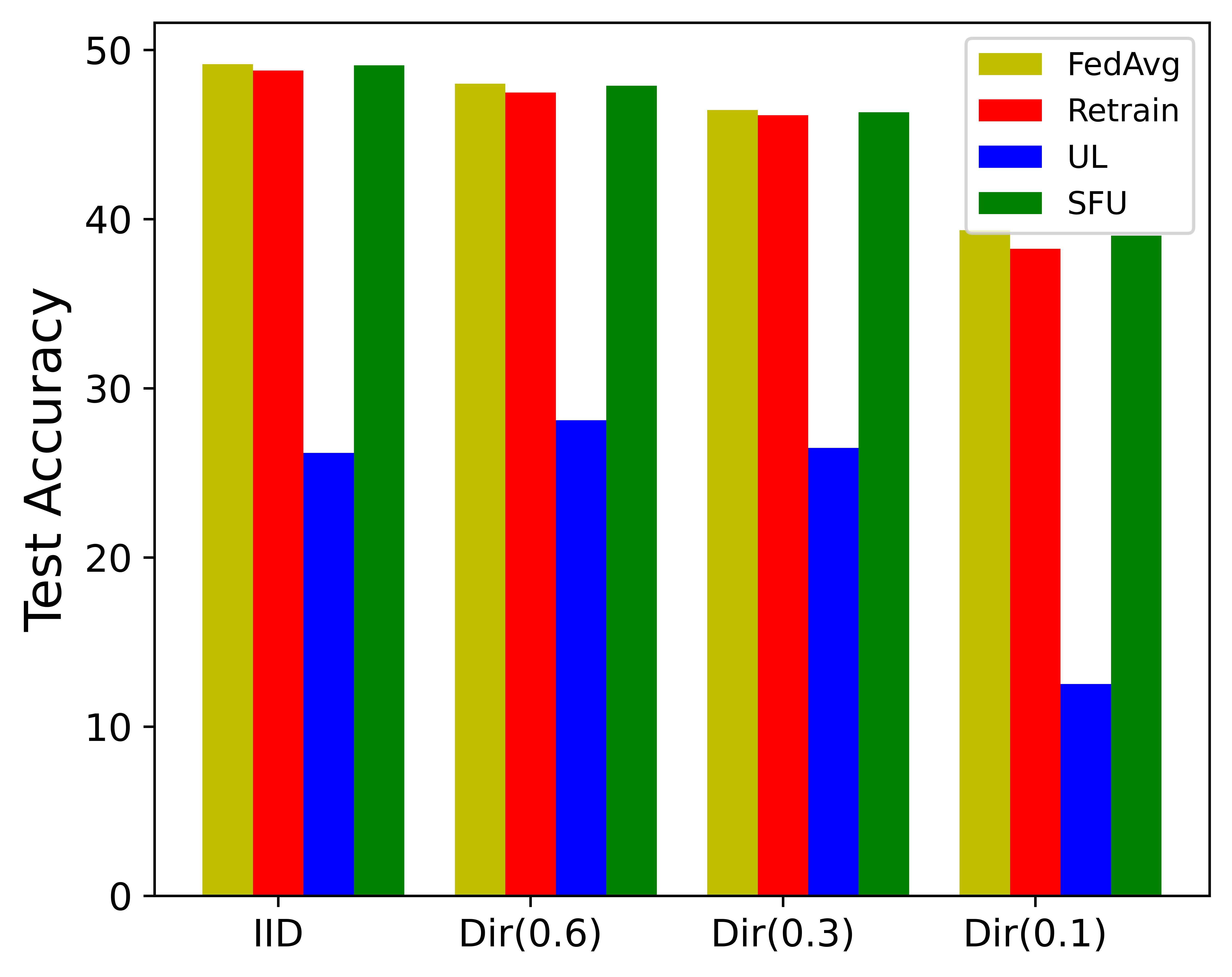}%
}
\hfill
\subfloat[CIFAR100]{\includegraphics[width=0.3\linewidth]{figure/cifar10_mlp3.png}}
\caption{Model accuracy after execution of SFU and other baselines under different degrees of data heterogeneity. The original global model is obtained by training with 10 clients, including one target client. }
\label{fig:he}
\vspace{-0.4cm}
\end{figure*}

\textbf{Models.}	We employ three neural network architectures in our experiments. 
\begin{itemize}
			\item {MLP:} This is a fully-connected neural network architecture with 2 hidden layers. The number of neurons in the layers is 200 and 100, respectively.

			\item {CNN:} This network architecture consists of 2 convolutional layers with 64 5 × 5 filters followed by 2 fully connected layers with 800 and 500 neurons and a      Relu	layer.
   
			\item {ResNet:} we use a smaller version of ResNet18 ~\cite{he2016deep}, with
three times fewer feature maps across all layers.
			
		\end{itemize}
We used two networks MLP and CNN for the MNIST dataset, while we used the above three network structures for the CIFAR10 and CIFAR100 datasets. We use PyTorch~\cite{oord2018representation} to implement those models.

\textbf{Evaluation metric.}\ 
 We use backdoor attacks in the target client’s updates to the global model as described before, so that we can intuitively investigate the unlearning effect based on the attack success rate of the unlearned global model. In Tab.~\ref{tabatk} and Tab.~\ref{tabacc}, we record the attack success rate as ``atk acc". A lower ``atk acc" represents a cleaner contribution removal from the target client. In our experiments, we implement the backdoor attack using a "pixel pattern" trigger of size 2 × 2 and change the label to "9". Because of the prediction error of the model, We can consider an error rate of less than 10\% as successful forgetting of the target client contribution. We also use the accuracy metric on the clean test data to measure the performance of the model after unlearning which is denoted ``test acc" in Tab.~\ref{tabatk} and Tab.~\ref{tabacc}. A high accuracy indicates that unlearning has little impact on the performance of the model.  Current unlearning methods usually adopt certain methods to recover the model accuracy after unlearning, so we divide our evaluation of the unlearning approach into two respects: the removal of specific client contributions and the recovery of model performance.

\begin{figure*}[ht]
\footnotesize
\centering
\setlength{\abovecaptionskip}{0.cm}
\subfloat[MNIST]{\includegraphics[width=0.3\linewidth]{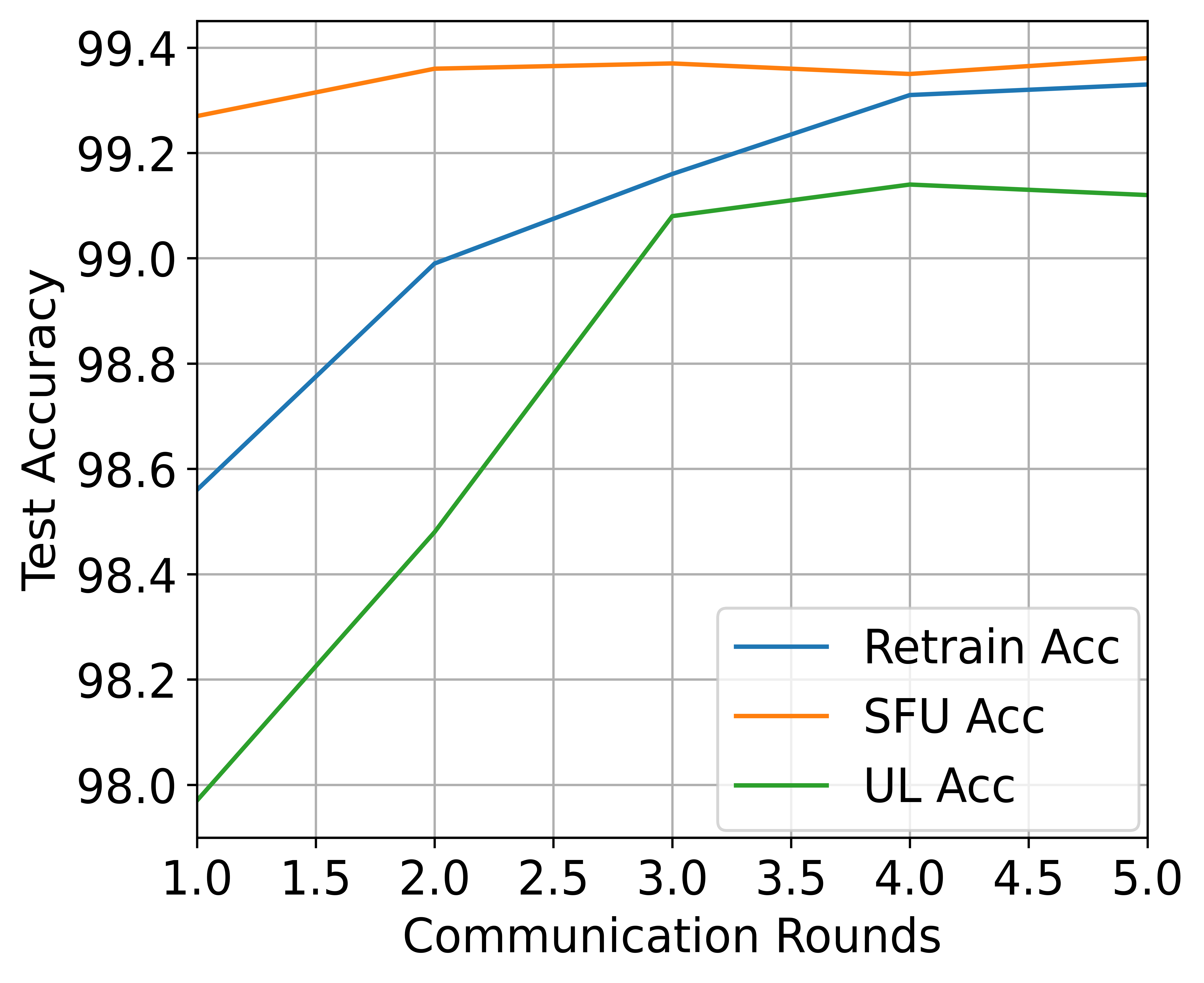}%
}
\hfill
\subfloat[CIFAR10]{\includegraphics[width=0.3\linewidth]{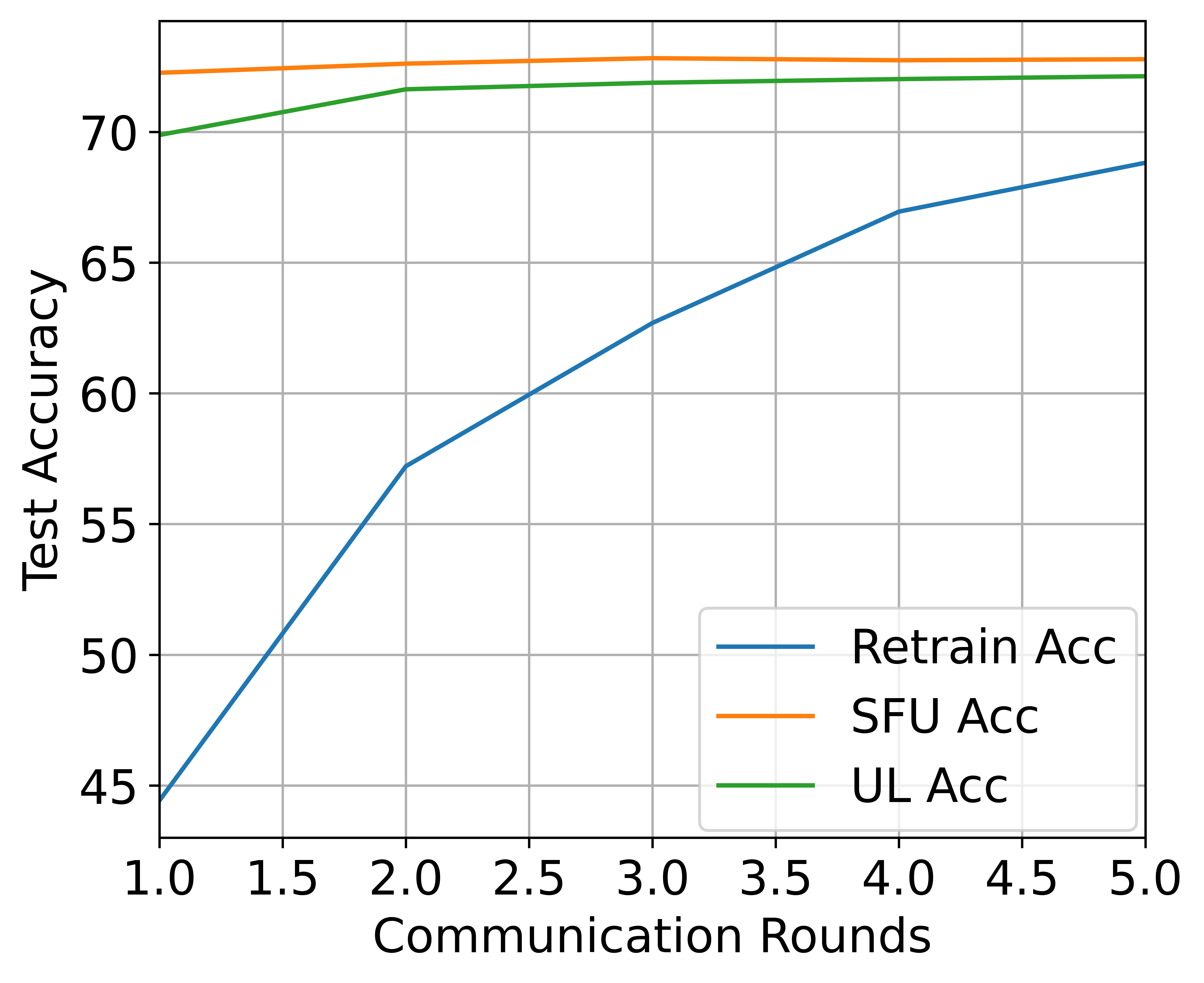}%
}
\hfill
\subfloat[CIFAR100]{\includegraphics[width=0.3\linewidth]{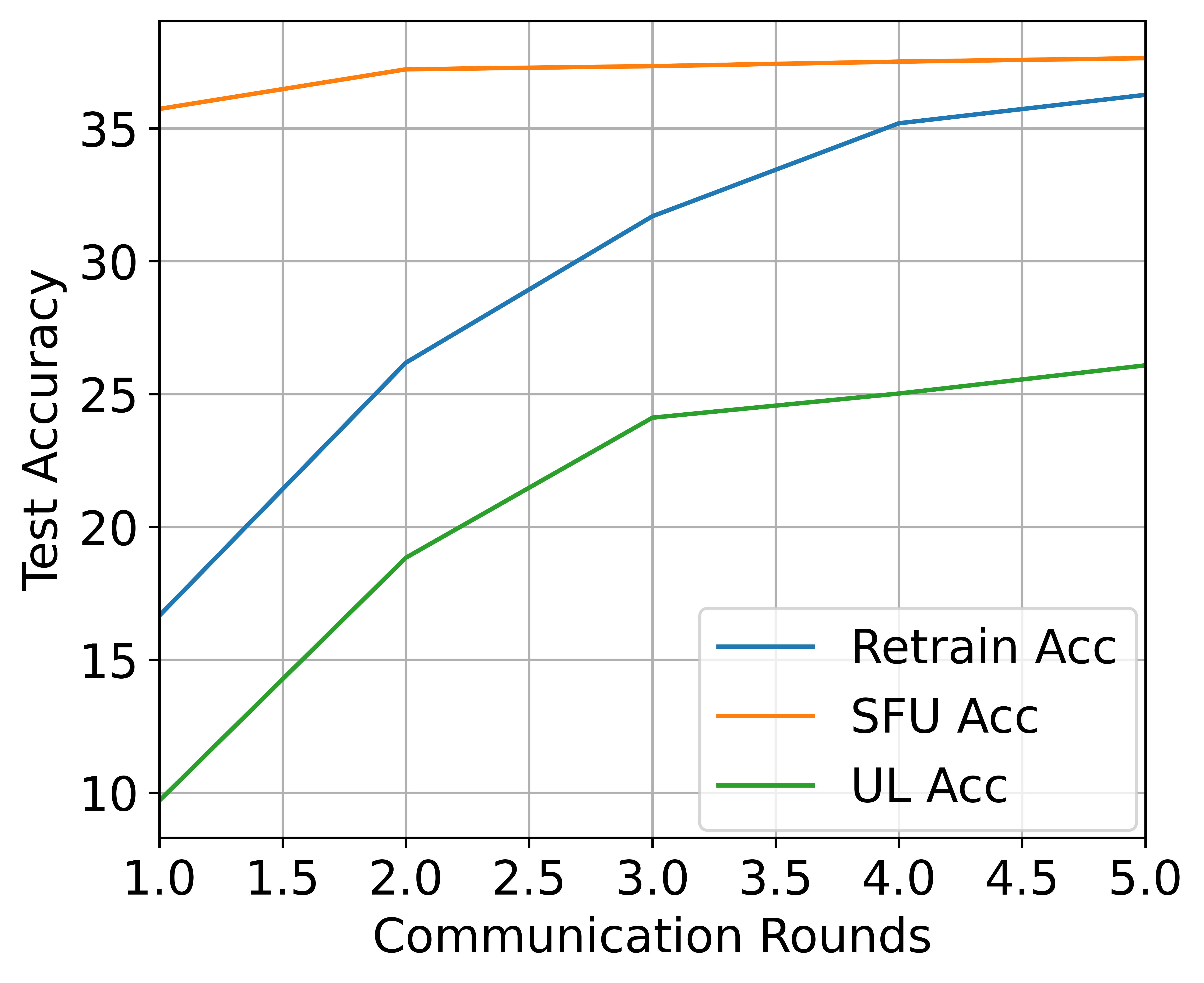}}
\caption{Convergence plots for SFU and other baselines in different datasets with CNN.}
\label{fig:re}
\vspace{-0.4cm}
\end{figure*}

\textbf{Implementation details.}\ We consider two scenarios: (i) We have ten clients with one target client, and all clients participate fully during each training round.  (ii) We have 100 clients with one target client and only 10\% of the clients participate during each training round.   In the experiment of removal of specific client contributions, We conduct unlearning experiments on the FL model after 100 rounds of training;
In the experiment of recovery of model performance, We start FL training with the stochastic model for full retraining. we start FL training on the unlearned local model without the involvement of the target client for SFU and GA. We use the knowledge of the public data on the server for distillation learning to recover the model accuracy for UL. 
Tab. ~\ref{tabacc} and Tab. ~\ref{tabatk}  only show the results of scenario (i). Results of scenario (ii) and other implementation details are shown in {\bf Appendix}~\ref{app} due to limited space.

\subsection{Results of Contribution Removal} \label{r1}
We conducted extensive experiments to determine the advantage of SFU in removing the contribution of a specific client in the global model. In addition, we demonstrate the robustness and superiority of SFU in different training degrees of the model and different data heterogeneity. All results are reported based on the global model after performing unlearning. The results show that SFU can effectively remove the contribution of target clients under the premise of ensuring  model accuracy, and has strong robustness to data heterogeneity and the training degree of the model.

\textbf{Efficient forgetting by SFU.}\
 Tab.~\ref{tabatk} reports the unlearning effects of SFU and the other baselines on different IID datasets and model architectures. The results show that SFU can efficiently complete the contribution removal of target clients. We observe that SFU successfully eliminates the contribution of target customers and achieves low precision for backdoor data. For example, SFU can reduce the backdoor attack success rate to or close to 0\% in all datasets. Other baselines also can achieve similar results. This shows that the current unlearning methods can effectively remove the contribution of the target client.

\textbf{High model accuracy for SFU.}\
The results in Tab. ~\ref{tabatk} also report the accuracy of the model on the clean test set data after unlearning for each baseline. The results show that SFU achieves the best results on various datasets and different model architectures. However, the UL method will cause great damage to the performance of the model, and the direct gradient ascent has a large probability of producing a low-accuracy model. For example, on the CNN architecture of the MNIST dataset, the accuracy of the UL model is only 42.76\%, which is 56.55\% lower than the accuracy of the original FL model 99.31\%. The accuracy of the model generated by GA is 92.33\%, which is 7\% lower than the 99.29\% accuracy of the model generated by SFU, which indicates that restricting the gradient in the orthogonal subspace of the input space can reduce the loss of the original model performance.

\textbf{Robustness of the original model maturity.}\
 Federated unlearning can occur at any stage of FL, so it is possible that the model does not converge when federated unlearning is performed. We define the maturity of the model in terms of the number of rounds of model training, where the more rounds the model is trained, the more mature it is. Fig.~\ref{fig:datadis} reports the effect of each baseline with unlearning at different rounds of the FL model training. The results show that SFU can ensure the performance of the model to the greatest extent at any round of model training, and the accuracy of SFU is similar to that of FL model training. UL is easy to remove in the early stage of model training. However, as the model continues to train, the integration degree of each client model increases, simply subtracting the historical weight of the client is easy to cause a significant decline in the accuracy of the model. For example, the UL accuracy in the MNIST dataset dropped from 97\% to 66\% as the model trained. GA has similar results to SFU, but still has lower accuracy than SFU, which indicates that  SFU  is robust to the original  model maturity.

\textbf{Robustness on heterogeneous data.}
 We test different unlearning algorithms on MNIST datasets with different degrees of heterogeneity, and ten clients are selected to form the FL system. Fig.~\ref{fig:he} shows that enhancing data distribution heterogeneity will reduce the effect of various unlearning algorithms. It is because the aggregation of the global model becomes more complex with the increase of model heterogeneity, which  makes it more challenging to separate the contributions of specific clients under the premise of ensuring model accuracy. However, the accuracy of SFU is close to that of the original FL model and the retrained model in all Settings, which indicates that SFU is robust to data heterogeneity.

 \subsection{Results of Model Accuracy Recovery}\label{r2}

 Below, we report the accuracy of the final model and the recovery efficiency of different methods. All results are reported based on the global model after unlearning.

\textbf{High final accuracy of the SFU.}
 Tab. \ref{tabacc} reports the model performance of each algorithm after 10 rounds of accuracy recovery training and after 100 rounds of full retraining. We find that almost all methods can achieve a low success rate of backdoor attacks except for GA. The reason why GA has a high success rate of backdoor attacks is that the algorithm after unlearning is still near the original model, and it is easy to converge to the original model with high accuracy and a high success rate of backdoor attacks. The final accuracy of SFU is the highest among all the methods, and sometimes it can even exceed the accuracy of retraining, which proves that updating in the subspace will produce a better-initialized model to achieve higher accuracy.

\textbf{High-speed precision recovery for SFU.}\ To compare the model accuracy recovery speed of SFU with each baseline, we calculate the accuracy and backdoor attack success rate of different methods in terms of the number of FL training rounds after unlearning, and the results are shown in Fig.~\ref{fig:re}. We observe that SFU is able to achieve high accuracy after one round of training, while other methods require 5 or even more rounds of training to achieve.  SFU is more efficient in terms of computation and communication cost on the retained clients than the baseline of retraining while achieving comparable  backdoor accuracy.

\section{Conclusion}
In this paper,  we propose a novel federated unlearning approach that can successfully eliminate the  contribution  of a specified client to the  global model, which also can minimize the model accuracy loss by performing a gradient ascent process within the subspace at any  stage of model training. Our approach only relies on the target client to be forgotten from the federation without the server or any  other client keeping track of its history of parameter updates. Our method also provides a differential privacy  method to protect the representation matrix information during training. We have used a backdoor attack to effectively evaluate the performance of the proposed method. Our experimental results demonstrate the efficiency and effectiveness of SFU.


\clearpage
\bibliography{example_paper}
\bibliographystyle{icml2023}

\clearpage
\newpage
\appendix
\onecolumn

\section{Appendix: More Experiment Results}\label{app}
We run experiments on the true world datasets of including MNIST, CIFAR10, 
 and CIFAR100.  We consider two scenarios: (i) We have ten clients with one target client, and all clients participate fully during each training round.  (ii) We have 100 clients with one target client and only 10\% of the clients participate during each training round.  We detailed describe the experiment settings and the experimental results of scenario (ii) in the following.

\subsection{Setups}

\paragraph{Dataset.}\
We adopt real-world datasets  including MNIST, CIFAR10, 
 and CIFAR100. MNIST\cite{xiao2017fashion} dataset contains 60,000 training data and 10,000 test data in 10 classes. Each data sample is a 28×28 grayscale image. CIFAR10 dataset contains 50,000 training data and 10,000 test data in 10 classes. Each data sample is a 3×32×32 color image. CIFAR100~\cite{krizhevsky2009learning} includes 50,000 training data and 10,000 test data in 100 classes as 500 training samples per class, as shown in Table~\ref{data}. For MNIST and CIFAR10/100, we normalize the pixel value within a specific mean and std value in our code, which are [0.5, 0.5, 0.5] for mean and [0.5, 0.5, 0.5] for std.
\begin{table*}[ht]	

\caption{The similarity between predicted and real data distribution}
	\label{data}
	\centering
	\resizebox{0.6\textwidth}{!}{
		\begin{tabular}{ccccc}
			\toprule
			Datasets & Training Data & Test Data  & Class & Size \\
			\midrule
   MNIST & 60000 & 10,000 & 10 & $28\times28$ \\
			CIFAR-10 & 50,000 & 10,000 & 10 & $3\times32\times32$ \\
			CIFAR-100 & 50,000 & 10,000 & 100 & $3\times32\times32$ \\

			\bottomrule
		\end{tabular}
	}
\end{table*}

\begin{figure*}[ht]
\centering
\subfloat[IID]{\includegraphics[width=0.4\textwidth]{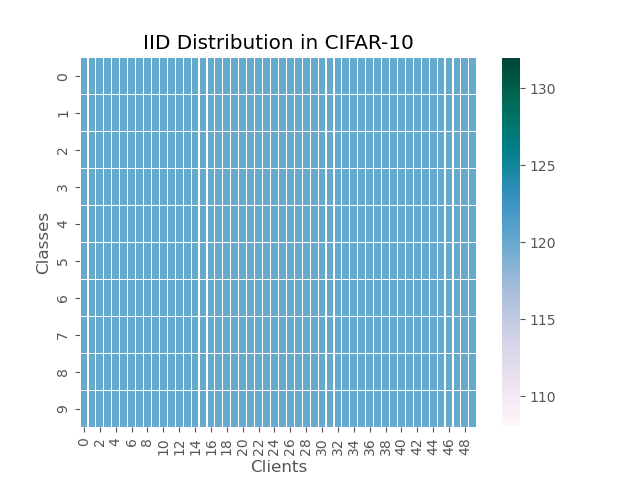}%
}
\hfill
\subfloat[Dir 0.3]{\includegraphics[width=0.4\textwidth]{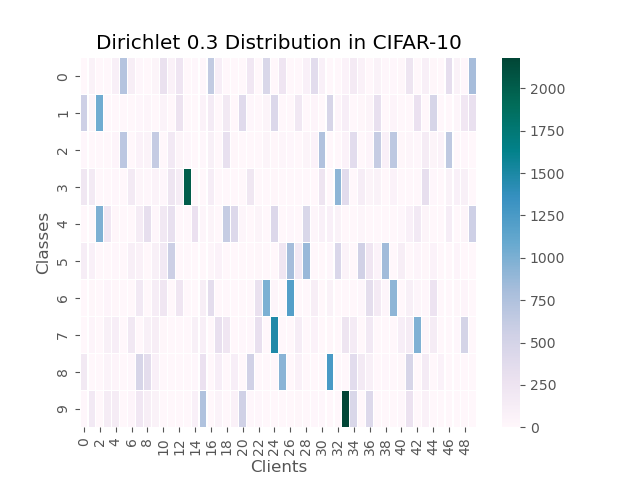}%
}

\caption{Heat maps for each client with  CIFAR10  dataset under different data partitions. The color bar denotes the number of data samples. Each rectangle represents the number of data samples of a specific class in a party. }
\label{fig:datadiss}
\end{figure*}

\paragraph{Dataset Partitions.}\
To fairly compare with the other baselines, we introduce the heterogeneity by splitting the total dataset by sampling the label ratios from the Dirichlet distribution. An additional parameter is used to control the level of  heterogeneity of the entire data partition. In order to visualize the distribution of heterogeneous data, we make the heat maps of the label distribution in different datasets, as shown in Fig.~\ref{fig:datadiss}.  It could be seen that for heterogeneity weight equals to 0.3 in Dirichlet distribution, about 10\% to 20\% of the categories dominate on each client, which is the blue block in Fig.~\ref{fig:datadis}.  The IID dataset is totally averaged in each client, which is the blue block in  Fig.~\ref{fig:datadis}.

\paragraph{Baselines.}\
\begin{itemize}
\item  Retraining the entire FL system without the target client being forgotten;

\item   Forgetting the target client based on the knowledge distillation ~\citet{wu2022federated}. This method subtracts  historical parameter updates  of the target client from the global model. Then, it uses the knowledge distillation method to remedy the skew of the unlearning model caused by the subtraction. We use ``UL" to denote this kind of algorithm in our experiments;

\item The global model is updated using gradient ascent information on the target client. To reduce the probability of producing a random model, we set the gradient clip norm when the global model is updated. We refer to this approach as ``GA" in our experiments.
\end{itemize}

 \paragraph{Implementation Details.}\
 In the experiment of removal of specific client contributions, We conduct unlearning experiments on the FL model after 100 rounds of training, where the hyperparameter Settings of each method are as follows:
For SFU, we set the learning rate as 0.01, epoch as 1, and mini-batch size as 64 for gradient ascent on the target client, and each client selects 10 samples to solve the local expression matrix. The random factor in differential privacy is generated by sampling from a uniform distribution of$[0.5-1]$. For SVD parameters we followed the setting of ~\citet{saha2021gradient}. For UL and GA, we set the same learning rate  and mini-batch as SFU, and the public data set of UL on the server  is formed by randomly sampling one-tenth of the total data.  
In the experiment of recovery of model performance, We start FL training with the stochastic model for full retraining. we start FL training on the unlearned local model without the involvement of the target client for SFU and GA. We use the knowledge of the public data on the server for distillation learning to recover the model accuracy for UL.

\subsection{Experimental results for scenario (ii).}

\begin{table*}
\centering
\footnotesize
\caption{Accuracy results after unlearning (100 clients).}
\label{tabatk2}
\begin{tabular}{|c|c|c|c|c|c|c|c|c|c|c|c|c|c|}
\hline

&&\multicolumn{2}{c|}{{FedAvg}}&\multicolumn{2}{c|}{{UL}}&\multicolumn{2}{c|}{{GA}}&\multicolumn{2}{c|}{{SFU}}\\
\hline

{Dataset}&{network}&{ test acc}&{atk acc}&{ test acc}&{atk acc}&{ test acc}&{atk acc}&{ test acc}&{atk acc}\\
\hline
\multirow{2}{*}{MNIST}  & {MLP}& 94.41 & 0.02 & 94.41 & 0.02 & 94.38 & 0.02 & \textbf{94.51} & 0.0
  \\  \cline{2-2}

&{CNN} &  98.47 & 0.0 & 98.47  & 0.0 & 98.26 & 0.0 & \textbf{98.44} & 0.0 \\  \cline{2-2}

\hline
\multirow{2}{*}{CIFAR10}  & {MLP} &  40.04 & 0.0 & 40.04 & 0.0 & 39.81 & 0.01 & \textbf{40.04} & 0.0   \\  \cline{2-2}

&{CNN} &  55.11 & 0.0  & \textbf{55.11} & 0.0 & 54.95 & 0.37 & {55.04}  & 0.0\\  \cline{2-2}
&
{ResNet} &  48.38 & 2.83 & \textbf{48.38} & 2.83 & {47.79} & 0.57 & {47.47} & 0.27  \\  \cline{2-2}
\hline
\multirow{2}{*}{CIFAR100}  & {MLP} & 13.5 & 0.0 & 13.5 & 0.0 & 13.47 & 0.0 & \textbf{13.51 }& 0.0  \\  \cline{2-2}

&{CNN} &  16.59 & 0.0 & 16.59 & 0.0 & 16.57 & 0.0 & \textbf{16.62} & 0.0   \\  \cline{2-2}
&
{ResNet} &  15.05 & 0.25 & 15.05 & 0.25 & \textbf{15.32} & 0.0& {14.93} & 0.0   \\  \cline{2-2}
\hline
\end{tabular}
\end{table*}

\begin{table*}
\centering
\footnotesize
\caption{Accuracy results after retraining (100 clients).}
\label{tabacc2}
\begin{tabular}{|c|c|c|c|c|c|c|c|c|c|c|c|c|c|}
\hline
&&\multicolumn{2}{c|}{{Retraining}}&\multicolumn{2}{c|}{{UL-Distillation}}&\multicolumn{2}{c|}{{GA-retraining}}&\multicolumn{2}{c|}{{SFU-retraining}}\\
\hline

{Dataset}&{network}&{ test acc}&{atk acc}&{ test acc}&{atk acc}&{ test acc}&{atk acc}&{ test acc}&{atk acc}\\
\hline
\multirow{2}{*}{MNIST}  & {MLP}& 94.44 & 0.02 & 94.44 & 0.02 & 94.43 & 0.02 & \textbf{94.98} & 0.0
  \\  \cline{2-2}

&{CNN} &  98.45 & 0.0 & 98.41  & 0.0 & 98.43 & 0.0 & \textbf{98.44} & 12.63  \\  \cline{2-2}

\hline
\multirow{2}{*}{CIFAR10}  & {MLP} &  40.13 & 0.0 & 40.13 & 0.0 & {39.95} & 20.36 & \textbf{41.49} & 0.10   \\  \cline{2-2}

&{CNN} &  55.11 & 0.00 & 55.27 & 0.0 & 55.36 & 0.0 & \textbf{56.13} & 0.0  \\  \cline{2-2}
&
{ResNet} &  47.33 & 3.51 & \textbf{50.93} & 2.52 & {48.36} & 1.63 & 49.93 & 0.94  \\  \cline{2-2}
\hline
\multirow{2}{*}{CIFAR100}  & {MLP} & 13.14 & 0.0 & \textbf{13.42} & 0.0 & {13.17} & 0.0 & {13.30 }& 0.0  \\  \cline{2-2}

&{CNN} &  16.97 & 0.0 & 16.57 & 0.0 & 16.57 & 0.0 & \textbf{17.51} & 0.0   \\  \cline{2-2}
&
{ResNet} &  15.39 & 0.12 & 15.01 & 0.25 & 14.92 & 0.14 & \textbf{16.09} & 0.09   \\  \cline{2-2}
\hline
\end{tabular}
\end{table*}

We compare the performance of SFU and other baselines on the setting of scenario (ii).Tab.~\ref{tabatk2}  and Tab.~\ref{tabacc2} show that the success rate of backdoor attacks in scenario (ii) are  lower than it in Tab.~\ref{tabatk}  and Tab.~\ref{tabacc}. When the client has a certain probability of being selected in the FL system, the target client has little contribution to the final model. However, the discussion about forgetting accuracy  is still consistent with the interpretation in Sec.~\ref{r1} and Sec.~\ref{r2}.


\end{document}